\documentclass{article}

\usepackage{arxiv}

\usepackage[utf8]{inputenc} 
\usepackage[T1]{fontenc}    
\usepackage{hyperref}       
\usepackage{url}            
\usepackage{booktabs}       
\usepackage{amsfonts}       
\usepackage{nicefrac}       
\usepackage{microtype}      
\usepackage{amsmath} 
\usepackage{cleveref}       
\usepackage{lipsum}         
\usepackage{graphicx}
\usepackage{natbib}
\usepackage{doi}
\usepackage{multirow}

\title{RL-Pruner: Structured Pruning Using Reinforcement Learning for CNN Compression and Acceleration}

\date{}

\author{
\bf Boyao Wang \thanks{Corresponding author} \quad Volodymyr Kindratenko\\
{University of Illinois Urbana-Champaign} \\
{\tt\small  boyaow2@illinois.edu, 
kindrtnk@illinois.edu} \\
}

\renewcommand{\shorttitle}{}

\hypersetup{
pdftitle={RL-Pruner: Structured Pruning Using Reinforcement Learning for CNN Compression and Acceleration},
pdfauthor={Boyao Wang, Volodymyr Kindratenko},
}

\begin{document}

\twocolumn[ 
  \begin{@twocolumnfalse}
  
\maketitle

\begin{abstract}
Convolutional Neural Networks (CNNs) have demonstrated exceptional performance in recent years. Compressing these models not only reduces storage requirements, making deployment to edge devices feasible, but also accelerates inference, thereby reducing latency and computational costs. Structured pruning, which removes filters at the layer level, directly modifies the model architecture. This approach achieves a more compact architecture while maintaining target accuracy, ensuring that the compressed model retains good compatibility and hardware efficiency. Our method is based on a key observation: filters in different layers of a neural network have varying importance to the model's performance. When the number of filters to prune is fixed, the optimal pruning distribution across different layers is uneven to minimize performance loss. Layers that are more sensitive to pruning should account for a smaller proportion of the pruning distribution. To leverage this insight, we propose RL-Pruner, which uses reinforcement learning to learn the optimal pruning distribution. RL-Pruner can automatically extract dependencies between filters in the input model and perform pruning, without requiring model-specific pruning implementations. We conducted experiments on models such as GoogleNet, ResNet, and MobileNet, comparing our approach to other structured pruning methods to validate its effectiveness. Our code is available at \url{https://github.com/Beryex/RLPruner-CNN}.
\end{abstract}
\vspace{0.35cm}
\end{@twocolumnfalse} 
] 

\footnotetext[1]{Corresponding author}

\section{Introduction}
\label{sec:Introduction}

Convolutional neural networks (CNNs) have demonstrated outstanding performance across a range of computer vision tasks, including image classification, detection, and segmentation \citep{computers12080151, bharadiya2023convolutional, oshea2015introductionconvolutionalneuralnetworks, GU2018354, 9451544}. As these architectures become wider and deeper, they gain an enhanced ability to extract complex features from input data. However, this increase in parameters leads to substantial computational costs, making inference both resource-intensive and slow. Moreover, there is a growing need to deploy CNNs on edge devices, which are often limited in computational power and memory \citep{9669989, stahl2021deeperthings}. Consequently, effective methods for compressing CNNs are increasingly in demand, aiming to reduce parameter counts and accelerate inference while maintaining the original model's performance. Such compression techniques are crucial for creating efficient and deployable CNNs that can meet the challenges of real-world applications.

Several techniques have been proposed to compress CNNs, most of which fall into one of four categories: structured and unstructured pruning, quantization, low-rank factorization, and knowledge distillation \citep{cheng2020surveymodelcompressionacceleration, 9043731}. Structured pruning \citep{Fang_2023_CVPR, 10330640, 10.1145/3005348} removes entire filters from neural networks and directly modifies the model's architecture, enabling both compression and realistic acceleration on standard hardware. Unstructured pruning \citep{Zhang_2018_ECCV, 9381660}, also known as weight pruning, removes specific unimportant elements from the weight matrix, which requires hardware or libraries that support sparse computation to achieve practical speedup. Low-rank factorization \citep{SWAMINATHAN2020185, Idelbayev_2020_CVPR} approximates the model's weight matrix by decomposing it into a product of low-rank matrices. Quantization \citep{Wu_2016_CVPR, gong2014compressingdeepconvolutionalnetworks, zhou2017incrementalnetworkquantizationlossless} reduces the bitwidth of the weight data in a model, achieving significant compression, but also requires hardware support to realize theoretical speedups for low-bit quantization, such as binary or ternary quantization. Knowledge distillation \citep{gou2021knowledge, Cho_2019_ICCV} transfers knowledge from a larger, more advanced teacher model to a smaller student model, which may not generate a new compressed architecture but helps recover the performance loss due to compression. In this paper, we primarily focus on structured pruning, though we hope the insights from our methods will also inform future work in other categories.

\begin{figure}[t!]
    \centering
    \includegraphics[width=\linewidth]{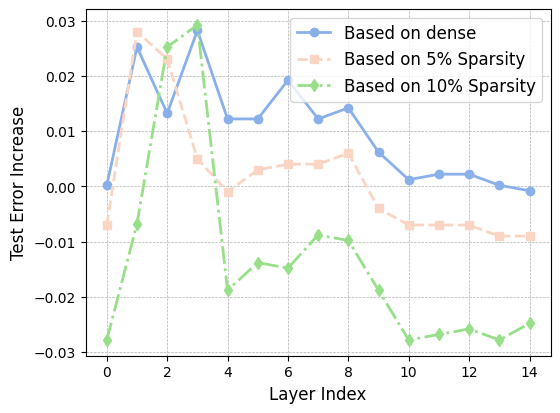}
    \caption{
         Increase in test error for a compressed VGG-16 model after pruning 10\% of filters from each individual layer. The pruning is conducted based on three different baselines: the dense model, a model with 5\% of filters pruned from all layers, and a model with 10\% of filters pruned from all layers. The results highlight that different layers have varying sensitivity to pruning, and this sensitivity changes dynamically throughout the pruning process. A negative increase indicates that performance improves.
    }
    \label{fig:layer_sensitivity}
    \vspace{-5mm}
\end{figure}

In structured pruning, while it is crucial to identify the least important filters to prune from weight matrices with minimal performance degradation, it is equally important to assess the importance of different layers to the model's performance and determine which layers should be pruned more or less. This involves learning the sparsity distribution across layers. As shown in Figure \ref{fig:layer_sensitivity}, the importance of different convolutional and linear layers to the overall model performance varies significantly. Moreover, after a certain extent of pruning, the relative importance among different layers also changes. To leverage this insight, we can assign higher filter sparsity to less important layers and lower sparsity to more important layers, adjusting the sparsity distribution dynamically throughout the pruning process to minimize the performance drop.

In this paper, we introduce a novel approach called RL-Pruner. Unlike other structured pruning methods that learn the sparsity distribution among layers during sparse training \citep{Liu_2017_ICCV, Ding_2021_ICCV, Fang_2023_CVPR}, RL-Pruner is a post-training structured pruning method that utilizes reinforcement learning with sampling to learn the optimal sparsity distribution across layers for several pruning steps. At each step, the current model architecture is defined as the state, while the pruning sparsity distribution acts as the policy. RL-Pruner adds Gaussian noise to the policy distribution to generate the real pruning sparsity distribution as an action, producing the corresponding compressed architecture as the next step's state. Each generation is treated as a sampling process. For each step, RL-Pruner maintains a replay buffer that stores the pruning sparsity distribution actions and their corresponding Q values, computed by the reward function. The reward function can be defined flexibly, such as based on the compressed model's test error, a combination of test error and parameter reduction ratio, or other criteria depending on the specific use case. After each pruning step, RL-Pruner updates the sparsity distribution policy to learn the optimal sparsity distribution. When computational resources allow, post-training stages are periodically applied after several pruning steps to recover any performance loss caused by pruning. We employ knowledge distillation in these post-training stages, with the original model acting as the teacher and the compressed model as the student.

RL-Pruner can automatically extract dependencies between filters in different layers of the input model by tracking tensor computations. Currently, our approach supports several popular CNN architectures, including residual connections, concatenation connections and skip connections. As a result, our method does not require model-specific pruning implementations and can perform pruning autonomously, enhancing the method's generality.

To validate the effectiveness of RL-Pruner, we apply the proposed method to several popular CNNs used for image classification, including VGGNet \citep{simonyan2014very}, ResNet \citep{he2016deep}, GoogLeNet \citep{szegedy2015going}, and MobileNet \citep{howard2017mobilenets}, using the CIFAR-10 and CIFAR-100 datasets \citep{krizhevsky2009learning}. According to the experimental results, RL-Pruner achieves 60\% channel sparsity for VGG-19 on CIFAR-100 and 40\% channel sparsity for GoogLeNet and MobileNetV3-Large on CIFAR-100, all with a performance drop of less than 1\%.

\section{Related Work}
\label{sec:Related Work}

\paragraph{Pruning} Pruning is one of the mainstream neural network compression methods and has made impressive progress \citep{Fang_2023_CVPR, 10330640, 10.1145/3005348, Zhang_2018_ECCV, 9381660}. Formally, given a neural network containing $L_c$ convolutional layers and $L_f$ fully connected layers, that is, $\mathcal{N} = \{W_l \in R^{N^{out}_l	\times N^{in}_l	\times K_l	\times K_l}, 1 \leq l \leq L_c\} \cup \{ W_l \in R^{N_{l+1}	\times N_l}, 1 \leq l \leq L_f \}$, where $N^{out}_l$, $N^{in}_l$ denotes the output and input channel number of the $l_{th}$ layer respectively. Each layer contains $N_{l+1}$ filters, with each filter having a shape of $N_l \times K_l \times K_l$ for convolutional layers and a shape of $N_l$ for fully connected layers. Unstructured pruning removes individual weights, resulting in sparsified filter matrices, which do not lead to inference speedups unless specialized hardware is available. In contrast, structured pruning operates at a larger granularity, removing entire filters and achieving both compression and realistic acceleration on standard hardware. Several pruning criteria have been proposed, including norm-based methods that prune weights or filters with the least norms \citep{li2017pruningfiltersefficientconvnets, he2018soft} and activation-based methods that use the activation map to identify unimportant weights or filters \citep{he2017channel, lin2020hrank, luo2017thinet}, and regularization-based methods that learn structured sparse networks by adding different sparsity regularizers \citep{hoefler2021sparsity, liu2017learning}. In this paper, we adapt the Taylor criterion, which combines the gradient of a weight, calculated from a calibration dataset, with its norm to determine the importance of weight filters.

\paragraph{Neural Architecture Search} Structured pruning removes filters and generates compressed subnetworks, which can be viewed as a form of neural architecture search (NAS). Since different layers in a model have varying importance to its overall performance, several works have focused on automatically assigning layer-wise sparsity. Among these, \citep{he2018amc, yu2021auto, yu2022topology} deploy reinforcement learning agents to automatically select the appropriate layer-wise pruning ratio by searching over the action space, i.e., the pruning ratio, without requiring manual sensitivity analysis. Gradient-based methods \citep{guo2020dmcp, ning2020dsa} modify the gradient update rule to make the optimization problem with sparsity constraints differentiable with respect to weights. Evolutionary methods \citep{liu2019metapruning, lin2020channel} use evolutionary algorithms to explore and search for sparse subnetworks. Our method primarily uses reinforcement learning to determine the optimal sparsity distribution across layers.

\paragraph{Knowledge Distillation} Pruning typically reduces a network's performance, so post-training is often employed to recover this loss. Knowledge distillation \citep{hinton2015distilling} transfers knowledge from a larger teacher model to a smaller student model. Response-based knowledge distillation allows the student model to mimic the predictions from the teacher model's final output layer by adding a distillation loss between their predictions to the optimization function. Feature-based knowledge distillation, on the other hand, enables the student model to learn intermediate layer representations from the teacher model. In our method, we adopt response-based knowledge distillation during the recovery post-training stage, as pruning alters the dimensions of intermediate layers, making knowledge transfer between these layers less straightforward.
\section{Method}
\label{sec:Method}

\begin{figure*}[!ht]
    \centering
    \includegraphics[width=\linewidth]{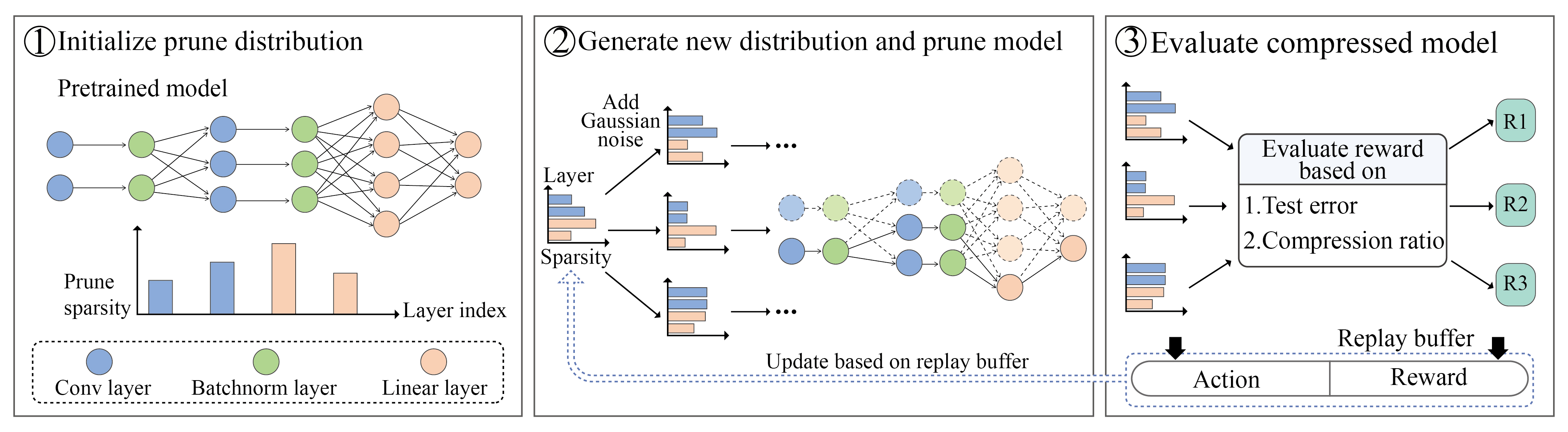}
    \caption{
         Illustration of RL-Pruner: We start by initializing a uniform pruning distribution based on the output channel density of each prunable layer. Next, we iteratively generate a new distribution, evaluate the corresponding compressed model to obtain a reward, and update the pruning distribution using the replay buffer.
    }
    \label{fig:method_description}
    \vspace{-5mm}
\end{figure*}

RL-Pruner first builds a dependency graph between layers in the model, then performs pruning in several steps. In each step: 1) A new pruning sparsity distribution is generated as an action based on the base distribution, which serves as the policy; 2) Each layer is pruned using the Taylor criterion according to the corresponding sparsity; 3) The compressed model is evaluated to obtain a reward, and the action and reward are stored in a replay buffer. After each step, the base distribution is updated according to the replay buffer, and if computational resources are sufficient, a post-training stage is applied to the compressed model using knowledge distillation, where the original model acts as the teacher. Figure \ref{fig:method_description} illustrates our method.

\subsection{Build Dependency Graph} 

\paragraph{Definition} Given a convolutional neural network $\mathcal{N}=\{W_l \in R^{N^{out}_{l} \times N^{in}_l\times K_l \times K_l}, 1 \leq l \leq L_c\} \cup \{W_l \in  R^{N^{out}_{l} \times N^{in}_l}, 1 \leq l \leq L_f\}$, we define convolutional and linear layers as prunable layers $\{\mathcal{L}^P_i, 1 \leq i \leq L_c + L_f\}$, while activation layers, and pooling layers are considered non-prunable layers. Batch normalization layers are also considered non-prunable, as they are only pruned if their preceding convolutional or linear layers are pruned. Formally, we establish the dependency between any two layers $\mathcal{L}_i, \mathcal{L}_j, i \leq j$ by,
\begin{align}
DG_{(i,j)}(k_{out}) = [k_{in}], \forall 1 \leq k_{out} \leq N^{out}_i
\end{align}
if a tensor flows from $\mathcal{L}_i$ to $\mathcal{L}_j$ with no intermediate layer between them. This means that if we prune a specific output channel $k_{out}$ in $\mathcal{L}_i$, the corresponding input channel $k_{in}$ in $\mathcal{L}_j$ must also be pruned. If there is no dependency between $\mathcal{L}_i$ and $\mathcal{L}_j$, we have:
\begin{align}
DG_{(i,j)}(k_{out}) = [-1], \forall 1 \leq k_{out} \leq N^{out}_i
\end{align}
Our goal is to build a dependency graph $DG$ among all layers, and record the relationships among their channel index mappings.

\paragraph{Build Dependency Graph} To achieve this, we fed the model an example input image and extract all layers $\{\mathcal{L}_i, 1 \leq i \leq L\}$ along with their input tensor $\{T^{in}_l, 1 \leq l \leq L\}$ and output tensor $\{T^{out}_l, 1 \leq l \leq L\}$, where $L$ is the number of all layers. We then build the dependency graph by matching the input and output tensors of each layer. A basic dependency occurs when the output tensor of one layer $\mathcal{L}_i$ is directly fed into layer $\mathcal{L}_j$, which implies for basic dependency
\begin{align}
DG_{(i,j)}(k_{out}) = [k_{out}], \forall 1 \leq k_{out} \leq N^{out}_i
\end{align}
We first iterate over $\{T^{in}\}$ and $\{T^{out}\}$ to check whether any two tensors match, in order to detect all the basic dependencies. Next, we check for any unused input tensor, e.g. $T^{in}_a$, which may be caused by tensor modifications (e.g., flattening) or special connections between layers (e.g., residual or concatenation connections). Flattening is commonly used between convolutional and linear layers. To detect the use of tensor flattening, we also flatten all output tensors and compare them to the unused input tensors to build the dependency. Let us denote the detected output tensor as $T^{out}_a$, which represents the tensor before flattening. To determine the mapping relationship after flattening, we need to check the output area $A_{out}$ of the preceding layer that produces $T^{out}_a$. For flattening, we have
\begin{align}
DG_{(i,j)}(k_{out})=[k_{out}*A_{out},(k_{out}+1)*A_{out}]
\end{align}
which indicates the output channel $k_{out}$ of $\mathcal{L}_i$ and the input channel from $k_{out}*A_{out}$ to $(k_{out}+1)*A_{out}$ must be pruned simultaneously. To detect concatenation connections, we iterate over $\{T^{out}\}$ to check whether there exists $T^{out}_b$ such that
\begin{align}
T^{in}_a = concat(T^{out}_b, T^{out}_j), 1 \leq j \leq L, j \neq b
\end{align}
This can be detected by slicing $T^{out}_b$ over $T^{in}_a$ such that $T^{in}_a[i_b, i_b+N^{out}_b]=T^{out}_b$. We can then build the dependency as 
\begin{align}
DG_{(b,a)}(k_{out}) = [i_b + k_{out}]
\end{align}
To detect residual connections, we iterate over $\{T^{out}\}$ to check whether there exists $T^{out}_b$, $T^{out}_c$ such that
\begin{align}
T^{in}_a = T^{out}_b + T^{out}_c, a \neq b, a \neq c, b \neq c
\end{align}
We can then build the dependency as 
\begin{align}
DG_{(b,a)}(k_{out}) = [k_{out}] \\
DG_{(c,a)}(k_{out}) = [k_{out}]
\end{align}
This also applies to the Squeeze-and-Excitation Module, where
\begin{align}
T^{in}_a = T^{out}_b * T^{out}_c, a \neq b, a \neq c, b \neq c
\end{align}
For residual connections and the Squeeze-and-Excitation Module, tensors from layers $\mathcal{L}_b$ and $\mathcal{L}_c$ are either added or multiplied and then passed to $\mathcal{L}_a$, meaning that pruning output channel $k^{out}$ of $\mathcal{L}_b$ also requires pruning output channel $k^{out}$ of $\mathcal{L}_c$ and vice versa. As a result, layers $\mathcal{L}_b$ and $\mathcal{L}_c$ must be pruned simultaneously. After constructing the dependency graph, we divide the layers that need to be pruned simultaneously into disjoint sets, ensuring that layers within the same set are pruned together.

\subsection{Assigning Layer-wise Sparsity}

After automatically extracting the layer dependencies and corresponding channel index mappings, we still need to determine how to distribute sparsity among layers for each pruning steps to minimize the performance drop caused by pruning while achieving the desired overall model sparsity. Let the pruning distribution $PD=\{\mathbf{p} \in [0, 1]^L \mid \sum_{i=1}^{L_c + L_f} p_i = 1\}$ represent the sparsity distribution among prunable layers, where we assign sparsity $p_i * S$ to the $i_{th}$ prunable layer, with $S$ being the target overall model sparsity. Since we can only prune output channels in integer quantities, the optimization problem of finding the optimal $PD$ is not differentiable.

To address this, we adopt a Monte Carlo sampling strategy \citep{rubinstein2016simulation} combined with the Q-learning algorithm \citep{watkins1992q} from reinforcement learning. Specifically, Monte Carlo sampling is used to explore different pruning distributions, while Q-learning updates the policy, represented by the base pruning distribution $PD$. Thus, each pruning step consists of multiple sampling stages followed by a pruning stage.

\paragraph{Sampling} We define each compressed model's architecture as the state and $PD$ as the policy. In each sampling, we add a Gaussian noise vector $\mathbf{z}$ to $PD$ to generate the real pruning action $a$:
\begin{gather}
\mathbf{z} = [z_1, z_2, \dots, z_L], z_i \sim \mathcal{N}(0, v), \forall i = 1, 2, \dots, L \\
a = PD + \mathbf{z}
\end{gather}
where $v$ is the variance of Gaussian noise that controls the exploration volume around $PD$. A larger $v$ indicates greater exploration, allowing the pruning policy to explore a wider range of potential architectures. Let $s$ denote the current model architecture, then, inspired by Bellman equation \citep{bellman1966dynamic}, we evaluate the Q value of each $a_i$ and the corresponding compressed model's architecture $s'_i$:
\begin{gather}
\label{eq:Q_value}
Q(s, a_i) = \mathbb{E} \left[r + \gamma \max_{a'_j} Q(s'_i, a'_j) \mid s, a_i \right], 1 \leq i,j \leq N_S \\
r = \mathcal{R}(s'_i)
\end{gather}
where $\gamma$ is the discount factor and $N_S$ is the sampling number at each time step. To achieve the sampling efficiency, we sample $T$ time steps in each sampling stage to approximate the expectation in equation \ref{eq:Q_value}:
{\small
\begin{gather}
Q(s, a_i) \approx \frac{1}{T} \sum_{t=1}^T \left[r_t + \gamma \max_{a'_{(t,j)}} Q(s'_{(t,i)}, a'_{(t,j)}) \right], 1 \leq i,j \leq N_S \\
r_t = \mathcal{R}(s'_{(t,i)})
\end{gather}}
\noindent
The reward function is determined by the compressed model's architecture, modeled by
\begin{gather}
\label{eq:reward}
\mathcal{R}(s) = T_e(s) + \alpha * C_F(s) + \beta * C_P(s)
\end{gather}
where $T_e, C_F, C_P$ represent the test accuracy, FLOPs compression ratio, and parameter count compression ratio of the input model, respectively. The hyperparameters $\alpha, \beta$ can be set to prioritize either accelerating inference or reducing parameter numbers, depending on the specific objective.

\paragraph{Update Policy} After each sampling stage, we store each action $a_i$ along with its Q value $Q(s, a_i)$ in a replay buffer $RB$, replacing the entry with the lowest Q value in $RB$. If the lowest Q value exceeds $Q(s, a_i)$, the corresponding sampling data is discarded. We then select an action $a^*$ from the replay buffer to update the policy $PD$. To balance exploration and exploitation, we use an $\epsilon$-greedy strategy: with probability $\epsilon$, we select $a^*$ randomly from $RB$. Otherwise, we select the action with the highest Q value:
\begin{gather}
a^* = \arg\max_{a_i} RB(s,a_i)
\end{gather}
As the model architecture is initially more redundant and becomes more compact through pruning, we use a high exploration $\epsilon$ value during the initial pruning steps, gradually reducing $\epsilon$ in the later steps. Several $\epsilon$ decay strategies are considered, including constant, linear, and cosine decay. We then adjust the policy distribution toward the selected action distribution by a step size $\lambda$. To ensure stability in the update process, we adopt the idea of proximal policy optimization (PPO) \citep{schulman2017proximalpolicyoptimizationalgorithms}, which constrains the change in the policy distribution:
\begin{gather}
PD_{new} = PD + \lambda * a^* \\
PD = clip(\frac{PD_{new}}{PD}, 1 - \delta, 1 + \delta) * PD
\end{gather}
where each element's change ratio is limited within $[1 - \delta, 1 + \delta]$. As a result, after several sampling stages, the policy $PD$ will converge toward an optimal pruning distribution across the layers.

\subsection{Layer Pruning}

Now that we have the dependency graph and the pruning sparsity for each layer, the next question is how to select which output channel within each layer to prune in order to minimize performance drop. For each convolutional layer $W_l \in R^{N^{out}_{l} \times N^{in}_l\times K_l \times K_l}$ and linear layer $R^{N^{out}_{l} \times N^{in}_l}$ that containing a total of $N^{out}_{l}$ output channels, each corresponds to a weight matrix, we use a calibration dataset extracted as a subset from the training dataset to help determine which weight matrix is less important. According to \citep{lecun1989optimal}, the importance of the $i_{th}$ weight matrix $W_i$ in layer $\mathcal{L}$ is given by:
\begin{align}
I_{W_i} &= |\Delta \mathcal{L}(\mathcal{D})| \\
&= |\mathcal{L}_{W_i}(\mathcal{D}) - \mathcal{L}_{W_i=0}(\mathcal{D})| \\
&= |{\frac{\partial \mathcal{L}^{\top}(\mathcal{D})}{\partial W_i} W_i}-\frac{1}{2} {W_i}^{\top} H W_i + \mathcal{O}\left(\| W_i \|^3\right) |
\end{align}
where $H = \frac{\partial \mathcal{L}^2(\mathcal{D})}{\partial W_i^2}$ is the Hessian matrix. Since computing the Hessian matrix requires $\mathcal{O} (N^2)$ computational resources, we omit it to accelerate the pruning process. We also disregard $\mathcal{O}\left(| W_i |^3\right)$, as its value is typically small compared to the first term. Thus, the estimated weight structure importance becomes:
\begin{align}
\hat{I}_{W_i} = |{\frac{\partial \mathcal{L}^{\top}(\mathcal{D})}{\partial W_i} W_i}|
\end{align}
After evaluating the importance of each output channel, we sort them and prune the least important ones based on the specified sparsity.

\begin{table*}[t]
  \centering
  \resizebox{\linewidth}{!}{
  \small
  \begin{tabular}{c l c | c | c | c}
      \toprule
      {\bf Architecture} & 
      \multirow{2}{*}{\bf Reward Strategy} & 
      \multicolumn{4}{c}{\footnotesize \bf Pruned Accuracy / FLOPs Ratio / Para. Num. Ratio at Different Sparsity Levels} \\
      \cmidrule(lr){1-1}  \cmidrule(lr){3-6} 
      \bf (Accuracy, FLOPs, Para. Num.) & & \bf 20\% & \bf 40\% & \bf 60\% & \bf 80\% \\
      \midrule
    \multirow{3}{*}{\shortstack{VGG-19\\(72.89, 418.63M, 39.33M)}}
    & Accuracy-Based  & 73.09 / 22.62 / 35.30 & 73.11 / 35.11 / 62.18 & 73.00 / 43.55 / 79.46 & 71.25 / 55.71 / 89.34 \\      
    & FLOPs-Based     & 73.05 / 26.09 / 36.58 & 72.65 / 39.40 / 63.12 & 72.50 / 47.33 / 81.04 & 71.51 / 53.96 / 90.38 \\  
    & Parameter-Based & 73.00 / 24.54 / 36.02 & 72.08 / 38.94 / 63.70 & 71.98 / 47.43 / 81.98 & 71.67 / 56.84 / 91.44 \\
    \midrule
    \multirow{3}{*}{\shortstack{ResNet-56\\(72.25, 127.93M, 0.86M)}}  
    & Accuracy-Based  & 70.79 / 51.01 / 37.67 & 68.33 / 73.36 / 66.19 & 59.65 / 87.02 / 89.32 & 37.74 / 95.51 / 97.77 \\      
    & FLOPs-Based     & 71.21 / 49.92 / 34.53 & 68.36 / 75.69 / 68.81 & 61.49 / 84.50 / 88.13 & 40.22 / 94.06 / 97.28 \\  
    & Parameter-Based & 71.41 / 51.05 / 37.28 & 68.46 / 75.86 / 70.80 & 60.68 / 87.19 / 88.32 & 40.02 / 95.65 / 97.69 \\
    \midrule
    \multirow{3}{*}{\shortstack{DenseNet-121\\(79.83, 907.93M, 7.05M)}}  
    & Accuracy-Based  & 78.60 / 38.59 / 36.22 & 77.08 / 67.26 / 64.60 & 73.33 / 85.72 / 82.60 & 69.10 / 95.26 / 94.50 \\      
    & FLOPs-Based     & 79.08 / 44.15 / 37.91 & 76.43 / 72.77 / 67.41 & 71.67 / 87.51 / 83.84 & 67.71 / 95.44 / 94.67 \\  
    & Parameter-Based & 78.44 / 39.76 / 39.10 & 76.21 / 68.32 / 67.25 & 73.85 / 85.98 / 84.14 & 69.73 / 94.95 / 95.01 \\
    \midrule
    \multirow{3}{*}{\shortstack{GoogLeNet\\(77.62, 535.66M, 6.40M)}} 
    & Accuracy-Based  & 77.61 / 31.78 / 34.09 & 76.78 / 53.88 / 60.55 & 74.17 / 72.65 / 81.05 & 69.96 / 89.85 / 95.23 \\      
    & FLOPs-Based     & 77.62 / 35.67 / 34.01 & 77.00 / 58.45 / 61.06 & 74.16 / 75.66 / 82.17 & 69.37 / 86.96 / 95.39 \\  
    & Parameter-Based & 77.64 / 32.21 / 34.49 & 76.47 / 55.57 / 60.75 & 73.85 / 71.10 / 81.18 & 69.47 / 88.73 / 94.88 \\
    \midrule
    \multirow{3}{*}{\shortstack{MobileNetV3-Large\\(56.18, 7.47M, 4.03M)}} 
    & Accuracy-Based  & 56.26 / 33.01 / 36.22 & 55.32 / 61.63 / 63.95 & 52.66 / 79.98 / 83.81 & 42.41 / 92.78 / 95.85 \\      
    & FLOPs-Based     & 56.26 / 33.01 / 36.22 & 55.32 / 61.63 / 63.95 & 53.28 / 80.17 / 83.74 & 41.65 / 93.14 / 95.91 \\  
    & Parameter-Based & 56.76 / 33.12 / 36.51 & 55.74 / 61.46 / 65.23 & 53.08 / 79.54 / 84.53 & 44.10 / 92.57 / 96.08 \\ 
     \bottomrule
  \end{tabular} 
  }
  \caption{Pruned accuracy, FLOPs compression ratio, and parameter reduction ratio on CIFAR-100 for various architectures under different reward strategies and sparsity configurations. The absolute FLOPs count and parameter number of the original architectures are listed in millions, while all other values are presented as percentages.}
  \vspace{-2mm}
  \label{tbl:results}
\end{table*}

\subsection{Post-train with Knowledge Distillation}

Given that the model is pruned multiple times to achieve the desired overall sparsity, and that pruning inevitably reduces performance, it is beneficial to periodically apply post-training after each pruning step to recover lost performance. Knowledge distillation \citep{hinton2015distilling} facilitates the transfer of knowledge from a more advanced teacher model to a smaller or compressed student model. To preserve the performance of the compressed models, we use the original model as the teacher and the corresponding compressed model as the student. Specifically, during post-training, we introduce a penalty term to the loss function, which measures the distance between the probability distributions of the teacher model, $p_t$, and the student model, $p_s$, based on the input image $x$. The modified loss function is:
\begin{align} 
Loss =& \tau * \text{Loss}(p_t(x),p_s(x)) \\
&+ (1 - \tau) * \text{Loss}(label, p_s(x))
\end{align}
where $\tau$ is a hyperparameter that controls the extent to which the student model learns from the teacher. As a result, the compressed model benefits from the knowledge of the original model, helping to maintain more of its performance.

\section{Experiment}
\label{sec:Experiment}

\subsection{Settings}

\paragraph{Setup} We build RL-Pruner from the scratch using Pytorch \citep{paszke2019pytorch}. Our experiments utilize an NVIDIA 4090D GPU with 24GB of memory and Intel(R) Xeon(R) Platinum 8481C CPU.

For implementing our methods, we conduct experiments using default hyperparameter values: noise variance $z = 0.04$, step size for policy update $\lambda = 0.1$, discount factor $\gamma = 0.9$, sample steps $T=1$ and sample number $N_S = 10$ for each sampling stage, with each pruning step containing 10 sampling stages. Proximal policy optimization is employed with $\delta=0.2$. We utilize three reward strategies in our experiments: accuracy-based, FLOPs-based, and parameter-based, corresponding to $(\alpha=0, \beta=0)$, $(\alpha=0.25, \beta=0)$ and $(\alpha=0, \beta=0.25)$ in equation \ref{eq:reward} respectively. For pruning each layer using the Taylor method, we extract 100 samples from the training dataset as the calibration dataset. A quick post-training is performed with a coefficient $\tau = 0.75$ for the loss between the teacher model and the student model. Furthermore, we switch the teacher model if a compressed model with higher accuracy is found during pruning. For the exploration strategy, we initialize the exploration parameter $\epsilon = 0.4$, which decreases in a cosine fashion over the first 10\% of pruning steps. Detailed hyperparameter analysis is provided \ref{sec:more_analysis}.

\paragraph{Models and Datasets} In this study, we focus on classification tasks, although our proposed methods are readily extendable to other tasks, such as segmentation. To demonstrate the general applicability of RL-Pruner for CNNs, we evaluate it on several widely used classification models, including VGG-19 \citep{simonyan2014very}, ResNet-56 \citep{he2016deep}, GoogLeNet \citep{szegedy2015going}, DenseNet121 \citep{huang2017densely}, and MobileNetV3-Large \citep{howard2017mobilenets}. These models encompass diverse architectural features such as residual connections, concatenation connections, and Squeeze-and-Excitation modules. For datasets, we utilize CIFAR-10 and CIFAR-100 \citep{krizhevsky2009learning} to assess performance. Floating point operations (FLOPs) determine computational complexity and inference speed, while the parameter count reflects memory usage during inference. In our experiments, we primarily use three criteria to evaluate the compressed model: test Top-1 accuracy, FLOPs compression ratio, and parameter count compression ratio. These ratios are computed as $1 - \frac{\text{FLOPs}(\text{pruned model})}{\text{FLOPs}(\text{base model})}$ and $1 - \frac{\text{Para. Num.}(\text{pruned model})}{\text{Para. Num.}(\text{base model})}$ respectively.

\paragraph{Baseline Setup} To validate the effectiveness of our method, we compare it against other structured pruning techniques, including DepGraph \citep{Fang_2023_CVPR} and GReg \citep{wang2020neural}. Additionally, we benchmark our method against other neural architecture search approaches, such as GNN-RL \citep{yu2022topology}. We evaluate all methods at three channel sparsity ratios—25\%, 50\%, and 75\%—to assess their performance under various levels of sparsity.

\begin{table*}[t]
  \centering
  \resizebox{\linewidth}{!}{
  \begin{tabular}{c l c | c | c}
      \toprule
      {\bf Architecture} & 
      \multirow{2}{*}{\bf Method} & 
      \multicolumn{3}{c}{\footnotesize \bf Pruned Accuracy / FLOPs Ratio / Para. Num. Ratio at Different Sparsity Levels} \\
      \cmidrule(lr){1-1}  \cmidrule(lr){3-5} 
      \bf (Accuracy, FLOPs, Para. Num.) & & \bf 25\% & \bf 50\% & \bf 75\%\\
      \midrule
    \multirow{4}{*}{\shortstack{VGG-19\\(72.89, 418.63M, 39.33M)}}
    & GReg & 72.09 / 29.17 / 44.34 & 71.55 / 52.72 / 71.31 & 53.41 / 82.35 / 89.24  \\
    & DepGraph  & 72.49 / 36.64 / 45.69 & \bf 72.69 / 43.97 / 71.64 & 72.05 / 75.38 / 91.32 \\
    & GNN-RL & 66.42 / 17.29 / 33.39 & 59.63 / 74.96 / 57.81 & N/A \\
    & RL-Pruner & \bf {}$^\dagger$73.13 / 26.50 / 42.83 & $^\dagger$72.65 / 39.29 / 72.05 & \bf 73.14 / 50.42 / 87.08 \\
    \midrule
    \multirow{4}{*}{\shortstack{ResNet-56\\(94.20, 127.93M, 0.86M)}}  
    & GReg & 93.36 / 52.56 / 40.17 & 89.31 / 79.28 / 74.17 & 84.35 / 95.37 / 93.72 \\
    & DepGraph & 93.69 / 48.47 / 45.83 & 92.44 / 74.01 / 77.90 & 83.85 / 92.86 / 94.36 \\
    & GNN-RL & 93.67 / 25.07 / 30.50 & \bf 92.45 / 84.40 / 53.97 & \bf 87.04 / 95.30 / 74.63 \\
    & RL-Pruner & \bf {}$^\dagger$93.71 / 53.44 / 49.36 & 91.76 / 80.09 / 84.25 & 86.63 / 91.39 / 95.97 \\
     \bottomrule
  \end{tabular} 
  }
  \caption{Pruned accuracy, FLOPs compression ratio, and parameter reduction ratio on for VGG-19 on CIFAR-100 and ResNet-56 on CIFAR-10 under different reward strategies and channel sparsity configurations. The absolute FLOPs count and parameter numbers for the original architectures are reported in millions, while all other values are expressed as percentages. The GNN-RL method prunes only the convolutional layers, while all other methods prune both convolutional and linear layers. RL-Pruner methods use an accuracy-based reward strategy. Each method undergoes a quick 100-epoch post-training phase without knowledge distillation. The highest accuracy at each sparsity level is highlighted in bold. {}$^\dagger$ indicates that post-training resulted in degraded performance, in which case we present the results without post-training.}
  \vspace{-2mm}
  \label{tbl:comparision}
\end{table*}

\subsection{Results}

\paragraph{Performance} We begin by evaluating our methods on popular CNN architectures using CIFAR-100. As shown in Table \ref{tbl:results}, we can compress VGG-19 to 60\% channel sparsity, and GoogleNet and MobileNetV3-Large to 40\% channel sparsity, with a performance drop of less than 1\%. This indicates that there is significant redundancy in these architectures that can be pruned without substantial performance loss, given the considerable reduction in FLOPs and parameter counts at these sparsity levels. We also note that ResNet-56's performance declines rapidly as pruning progresses. This may be due to ResNet-56's many layers with relatively few parameters, resulting in a low number of channels per layer. Consequently, the optimal sparsity distribution learned by our algorithm is easily rounded to zero during pruning, which limits its ability to represent each layer's relative importance and leads to a performance drop.

\paragraph{Structured Pruning Methods Comparison} Following previous work \citep{Fang_2023_CVPR}, we compare our method to other structured pruning and neural architecture search methods using two popular architectures: VGG-19 on CIFAR-100 and ResNet-56 on CIFAR-10. As shown in Table \ref{tbl:comparision}, our method achieves higher performance for VGG-19 across all three tested levels of channel sparsity. For the VGG-19 architecture, our method identifies the linear layers as less important and allocates more channel sparsity to them, resulting in a significantly lower FLOPs compression ratio at 75\% sparsity. Note that the GNN-RL method does not support pruning VGG-19 at 75\% sparsity in our reproduction. For ResNet-56, our method achieves higher performance at 25\% sparsity, along with greater FLOPs and parameter compression ratios. However, we observe suboptimal results at 50\% and 75\% sparsity, likely due to the low number of channels per layer in ResNet-56.

\subsection{More Analysis}
\label{sec:more_analysis}

In this section, we explore the effects of various methods employed during the pruning process, such as reward strategies and exploration ($\epsilon$). To minimize errors introduced by complex architectures, we conduct our experiments on the simpler VGG-19 architecture to evaluate these methods.

\paragraph{Effects of Reward Strategy}

\begin{figure}[t!]
    \centering
    \includegraphics[width=\linewidth]{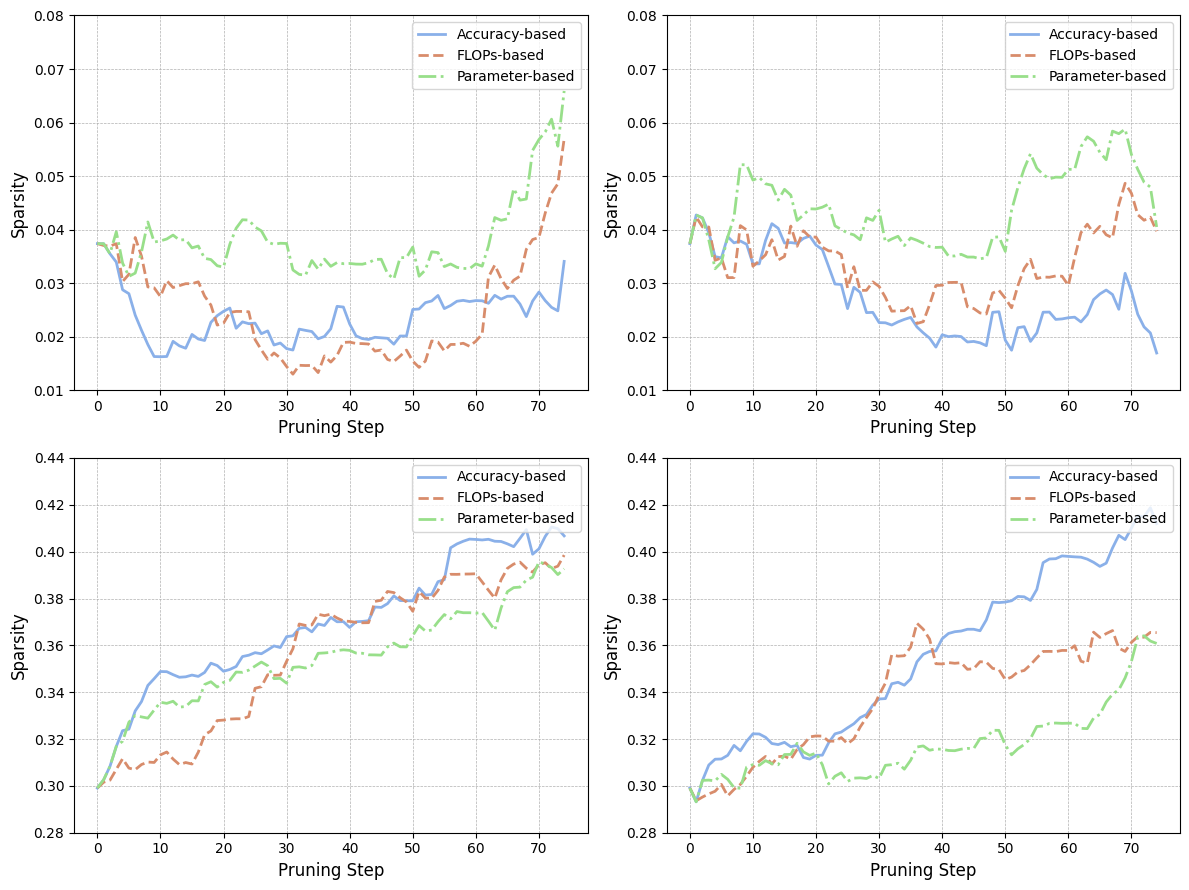}
    \caption{
         Sparsity changes across different layers during the pruning of VGG-19 on CIFAR-100 over 75 pruning steps with various reward strategies. The first row (from left to right) shows the sparsity changes in the $14^\text{th}$ and $16^\text{th}$ convolutional layers, while the second row (from left to right) depicts the sparsity changes in the $1^\text{st}$ and $2^\text{nd}$ linear layers of the architecture.
    }
    \label{fig:layer_sparsity}
    \vspace{-5mm}
\end{figure}

We first evaluate how different reward strategies influence pruning results. We track the sparsity changes of specific layers in VGG-19 over 75 pruning steps. As shown in Figure \ref{fig:layer_sparsity}, the parameter-based reward strategy tends to prune more filters in the convolutional layers, while the accuracy-based reward strategy favors pruning more neurons in the linear layers. It is challenging to establish a definitive principle linking the reward strategy to which layers are pruned more. Instead, our method identifies the optimal architecture based on the given reward strategy.

\paragraph{Exploration $\epsilon$ Choice}

\begin{table}[t]
  \centering
  \resizebox{\linewidth}{!}{
  \small
  \begin{tabular}{c l | c c c}
      \toprule
      \multirow{2}{*}{\bf Exploration $\epsilon$} & \multirow{2}{*}{\bf Decay Strategy} & \multicolumn{3}{c}{\footnotesize \bf Acc. at Different Sparsity Levels} \\
      \cmidrule(lr){3-5} 
       & & \bf 25\% & \bf 50\% & \bf 75\% \\
      \midrule
      0 & constant & 72.94 & 72.33 & 72.00 \\
      \midrule
      \multirow{3}{*}{0.2} 
      & constant  & 72.71 & 72.24 & \bf 72.53 \\
      & linear    & {}$^\dagger$ 73.10 & {}$^\dagger$ 72.32 & {}$^\dagger$ 72.03 \\
      & cosine    & {}$^\dagger$ 73.10 & {}$^\dagger$ 72.32 & {}$^\dagger$ 72.03 \\
      \midrule
      \multirow{3}{*}{0.4} 
      & constant  & 72.98 & 72.51 & 72.19 \\
      & linear    & 73.10 & 72.32 & 72.03 \\
      & cosine    & \bf 73.46 & \bf 72.84 & 71.60 \\
      \bottomrule
  \end{tabular} 
  }
  \vspace{1mm}
  \caption{Comparison of different exploration $\epsilon$ choices. The highest accuracy at each sparsity level is highlighted in bold. {}$^\dagger$: the results are the same caused by randomness.}
  \vspace{-2mm}
  \label{tbl:comparison_epsilon}
\end{table}

We evaluate different initial values of exploration $\epsilon$ and reduce it to 0 over the first 10\% of pruning steps using various decay strategies for VGG-19 on CIFAR-100. As shown in Table \ref{tbl:comparison_epsilon}, adopting an exploration strategy can lead to higher performance by exploring a broader range of architectures. However, it may also occasionally converge to a suboptimal architecture, resulting in reduced performance.

\paragraph{Compared to Scratch Training}

\begin{table}[t]
  \centering
  \resizebox{\linewidth}{!}{
  \small
  \begin{tabular}{c c | c c c}
      \toprule
      \bf Model / Dataset & \bf Sparsity & \bf Compressed  & \bf Trained & \bf $\Delta$ Acc. \\
      \midrule
      \multirow{3}{*}{\shortstack{VGG19\\CIFAR100}} 
      & 40\%    & 73.11 & 72.60 & +0.51 \\
      & 60\%    & 73.00 & 72.10 & +0.90 \\
      & 80\%    & 71.25 & 69.50 & +1.75 \\
      \bottomrule
  \end{tabular} 
  }
  \vspace{1mm}
  \caption{Comparison to models training from the scratch}
  \vspace{-2mm}
  \label{tbl:compared_to_scratch_training}
\end{table} 

We then validate the distillation effects of our method. Specifically, we train a compressed version of the VGG-19 model, where the architecture is initialized to match the compressed model, from scratch, and compare its performance to that of the compressed model. As shown in Table \ref{tbl:compared_to_scratch_training}, training the model initialized with the compressed architecture from scratch does not achieve the same performance as the compressed model. Moreover, the performance gap increases as channel sparsity grows. This suggests that pruning combined with post-training not only accelerates inference and reduces parameter count but also guides the model to learn features more efficiently, leading to higher knowledge density in the neural network—something that cannot be achieved through training alone. This phenomenon becomes more pronounced as sparsity increases.

\paragraph{Computational Cost}

Since our method constructs dependency graphs by tracking and comparing tensors, building the dependency graph for complex architectures (e.g., ResNet and DenseNet) takes several minutes, while for simpler architectures (e.g., VGG), it takes less than a minute. Given that our approach involves dozens of pruning steps, with each step comprising multiple sampling stages, the entire pruning process takes several hours under our experimental settings. There is potential to accelerate the sampling process by parallelizing it, as each sampling operation within a sampling stage is independent.
\section{Conclusion}
\label{sec:Conclusion}

In this paper, we present RL-Pruner, a structured pruning method that learns the optimal sparsity distribution across layers and supports general pruning without model-specific modifications. We hope our approach, which recognizes that each layer has a different relative importance to the model's performance, will influence future work in neural network compression, including unstructured pruning and quantization.

\bibliographystyle{unsrtnat}
\bibliography{references}  






\end{document}